\documentclass[10pt,conference]{IEEEtran}
\IEEEoverridecommandlockouts
\usepackage{amssymb}
\usepackage{amsmath,amsfonts}
\usepackage{array}
\usepackage[caption=false,font=normalsize,labelfont=sf,textfont=sf]{subfig}
\usepackage{textcomp}
\usepackage{stfloats}
\usepackage{url}
\usepackage{verbatim}
\usepackage{graphicx}
\usepackage{cite}
\usepackage{amsmath}
\usepackage{autobreak}
\usepackage{algorithmic}
\usepackage{booktabs}
\usepackage{bm}
\usepackage{float}
\usepackage{soul}
\usepackage{xcolor}
\usepackage{multirow}
\usepackage{graphicx}
\usepackage{tikz}
\usepackage{chngcntr}

\usepackage[linesnumbered,boxed,ruled,commentsnumbered]{algorithm2e}
\usepackage{textcomp}
\usepackage{threeparttable}

\def\BibTeX{{\rm B\kern-.05em{\sc i\kern-.025em b}\kern-.08em
    T\kern-.1667em\lower.7ex\hbox{E}\kern-.125emX}}
\begin{document}

\title{A Dual Large Language Models Architecture with Herald Guided Prompts for Parallel Fine Grained Traffic Signal Control
% \thanks{Identify applicable funding agency here. If none, delete this.}
}

\author{
  \IEEEauthorblockN{
    Qing Guo\IEEEauthorrefmark{1},
    Xinhang Li\IEEEauthorrefmark{1},
    Junyu Chen\IEEEauthorrefmark{1},
    Zheng Guo\IEEEauthorrefmark{1},\\
    Xiaocong Li\IEEEauthorrefmark{3},
    Lin Zhang\IEEEauthorrefmark{1}\textsuperscript{,}\IEEEauthorrefmark{2},
    Lei Li\IEEEauthorrefmark{1}
  }
  \IEEEauthorblockA{\IEEEauthorrefmark{1} School of Artificial Intelligence, Beijing University of Posts and Telecommunications, Beijing, China}
  \IEEEauthorblockA{\IEEEauthorrefmark{2} Beijing Big Data Center, Beijing, China}
  \IEEEauthorblockA{\IEEEauthorrefmark{3} Eastern Institute of Technology, Ningbo, China}
  % \IEEEauthorblockA{
  %   Emails: \{youshen,lixinhang,junyuchen,gzheng,zhanglin,leili\}@bupt.edu.cn;\
  %   li\_xiaocong@simtech.a-star.edu.sg;\ 
  %   eleaam@nus.edu.sg
  % }
  \thanks{Corresponding author: Lei Li (leili@bupt.edu.cn).}
}

\maketitle

\begin{abstract}
    Leveraging large language models (LLMs) in traffic signal control (TSC) improves optimization efficiency and interpretability compared to traditional reinforcement learning (RL) methods. However, existing LLM-based approaches are limited by fixed time signal durations and are prone to hallucination errors, while RL methods lack robustness in signal timing decisions and suffer from poor generalization. To address these challenges, this paper proposes HeraldLight, a dual LLMs architecture enhanced by Herald guided prompts. The Herald Module extracts contextual information and forecasts queue lengths for each traffic phase based on real-time conditions. The first LLM, LLM-Agent, uses these forecasts to make fine grained traffic signal control, while the second LLM, LLM-Critic, refines LLM-Agent’s outputs, correcting errors and hallucinations. These refined outputs are used for score-based fine-tuning to improve accuracy and robustness. Simulation experiments using CityFlow on real world datasets covering 224 intersections in Jinan (12), Hangzhou (16), and New York (196) demonstrate that HeraldLight outperforms state of the art baselines, achieving a 20.03\% reduction in average travel time across all scenarios and a 10.74\% reduction in average queue length on the Jinan and Hangzhou scenarios. The source code is available on GitHub: https://github.com/BUPT-ANTlab/HeraldLight.
\end{abstract}


\section{Introduction}

Advancements in artificial intelligence (AI) and sensor technologies are rapidly propelling the evolution of Intelligent Transportation Systems (ITS), offering significant potential to improve traffic efficiency and reduce congestion. Among these developments, Traffic Signal Control (TSC) has become a crucial solution for optimizing traffic efficiency \cite{guo2023urban}. Most current approaches rely on fixed timing strategies, which lack adaptability to varying traffic conditions, limiting system efficiency. In contrast, dynamic timing requires managing additional decision variables and precise control to respond to real-time traffic fluctuations. Addressing such challenges is essential for improving urban traffic management and optimizing the performance of transportation systems.%\cite{ zhao2022learning}.

Reinforcement learning (RL) has been extensively utilized in traffic signal control (TSC) as an effective approach for optimizing traffic flow and alleviating congestion. Xu et al. proposed HiLight, a hierarchical RL framework designed for short-term traffic optimization \cite{xu2021hierarchically}. Gu et al. introduced $ \pi $-Light, an interpretable RL model tailored for resource-limited settings \cite{gu2024pi}. Liang et al. proposed OAM for multi-intersection control \cite{liang2022oam}. RL methods has demonstrated effectiveness in decentralized multi-agent systems. However, most RL methods adopt a fixed timing strategy that limits flexibility, while RL methods employing dynamic timing often provide limited action-level justification and exhibit weak generalization.

Recent advancements have shown that large language models can address interpretability issues in RL methods \cite{lai2023large}. By leveraging the reasoning capabilities of LLMs, traffic signal control systems can benefit from transparent and consistent decision-making \cite{10763740, 10763542}. Approaches like PromptGAT have been used to adapt RL-trained policies to real-world traffic conditions \cite{da2024prompt}, while Open-TI facilitated coordinated traffic analysis and policy training \cite{da2024open}. Integration frameworks such as iLLM-TSC combined RL with LLMs to manage incomplete state information for real-time adjustments \cite{pang2024illm}. However, challenges persist in adapting to dynamic traffic signal timing, particularly in complex urban environments. Additionally, the occurrence of hallucinations of LLMs may lead to inaccurate decisions\cite{10.1145/3703155}, thereby undermining reliability. Addressing these limitations is necessary to make LLMs based TSC systems more adaptive and efficient for urban traffic management.

Existing studies have attempted to tackle the high dimensionality of the action space in dynamic signal timing. Wang et al. proposed UniTSA for single-intersection control and reported fixed time variants (Fix-30/Fix-40)\cite{10535743}. Zhang et al. introduced DynamicLight, a two-stage dynamic timing method that first selects the active phase and then sets its duration\cite{zhang2024dynamiclighttwostagedynamictraffic}. Kim et al. developed a deep RL strategy for prioritized phase split optimization, dynamically adjusting signal durations to improve network efficiency \cite{kim2023prioritized}. Above methods improve the accuracy of action selection for TSC. However, the limited interpretability and weak generalization of existing methods constrain the broader adoption of dynamic signal timing.

% Most RL and LLM based TSC methods rely on fixed timing strategies with predetermined signal durations. In contrast, some approaches focus on dynamic signal timing, adjusting durations in real-time to respond to traffic fluctuations. For instance, Mo et al. proposed CVLight, a decentralized RL-based method for multi-intersection control that adjusts phase durations based on real-time data \cite{MO2022103728}. Kim et al. developed a deep RL strategy for prioritized phase split optimization, dynamically adjusting signal durations to improve network efficiency \cite{kim2023prioritized}. Zhang et al. introduced an off-policy Nash deep Q-network algorithm for large-scale signal control with dynamic phase adjustments \cite{zhang2023large}. These methods mark significant progress in dynamic signal timing strategies. However, these approaches still face challenges, such as insufficient flexible control over signal timing, which limits effective coordination and adaptability to dynamic traffic conditions. Furthermore, the lack of interpretability in decision-making processes undermines scalability and reliability.

\begin{figure*}[t]
\centering
\includegraphics[width=\textwidth]{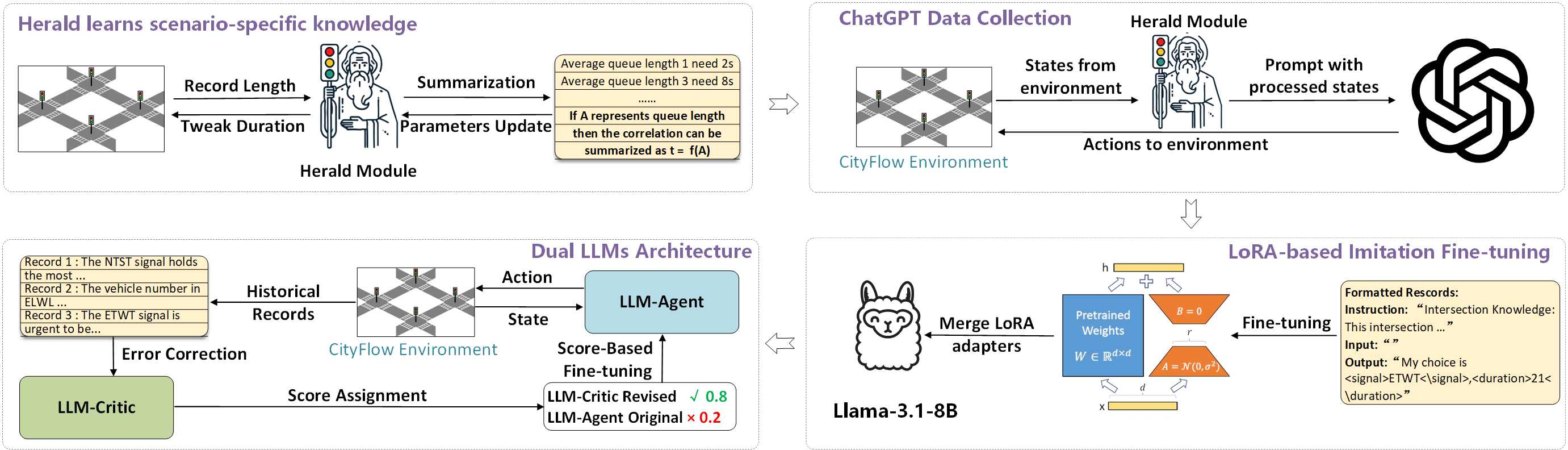}
\caption{Overview of HeraldLight. Herald Module learns and updates; ChatGPT assisted with Herald Module curates imitation data; Llama-3.1-8B is LoRA-finetuned to instantiate the LLM-Agent; an LLM-Critic (ChatGPT) evaluates and revises decisions; scored corrections drive iterative, score-based fine-tuning.}
\label{fig:framework}
\end{figure*}

This paper presents HeraldLight, a dual LLMs architecture driven by Herald guided prompts. The LLM-Agent, enhanced via LoRA-based imitation fine-tuning, infers the active signal phase and duration from real-time traffic conditions. The LLM-Critic, informed by Herald, evaluates and revises LLM-Agent’s outputs to mitigate hallucinations and improve consistency and reliability. An overview is provided in Fig.~\ref{fig:framework}, and the contributions are summarized as follows:

\begin{itemize}
    \item 
    The Herald Module evaluates intersection traffic flow by extracting environmental features and projecting queue states up to 40 seconds ahead. By precisely predicting queue lengths and accurately modeling the relationship among environmental factors, signal phase durations, and queue lengths, it enables dynamic traffic signal control with second-level fine grained adjustments.
    \item 
    This paper proposes a collaborative agent-critic architecture that integrates two LLMs to advance environmental analysis and reasoning in TSC. Through iterative interactions between the LLM-Agent and the LLM-Critic, the architecture progressively mitigates hallucination and systematically enhances the reasoning capabilities required for dynamic TSC strategies.
    \item
    All experiments are conducted in the CityFlow simulator using three open-source real-world datasets. Jinan (1--3), Hangzhou (1--2), and New York (1--2). Compared with state-of-the-art method, the proposed approach achieves a 20.03\% reduction in Average Travel Time (ATT) across all scenarios and a mean 10.74\% reduction in Average Queue Length (AQL) on the Jinan and Hangzhou networks.

\end{itemize}

The rest of this paper is organized as follows: Section 2 describes the intersection modeling. Section 3 presents the
composition of the Herald Module. Section 4 outlines HeraldLight framework including prompt design, imitation
fine-tuning and Dual LLMs architecture. Section 5 presents the experimental setup and results. Section 6 concludes with a summary and future directions.

\section{Intersection Modeling}\label{sectionII}

\subsection{Lanes, Phases, and Duration}

\textbf{Lanes.}
For each direction \(d\in\{\mathrm{N},\mathrm{S},\mathrm{E},\mathrm{W}\}\), incoming lanes are grouped by movement
\(L^{\mathrm{in}}_{d}=\{L^{l}_{d},L^{s}_{d},L^{r}_{d}\}\) (left/straight/right), with corresponding outgoing lanes \(L^{\mathrm{out}}_{d}\).
Right-turn lanes are treated as permissive (unsignalized).

\textbf{Phases.}
Let \(P=\{P_{1},P_{2},P_{3},P_{4}\}\).
Each phase controls two lanes (two opposing approaches):
\(P_{1},P_{2}\) serve the N--S approaches (straight/left), and \(P_{3},P_{4}\) serve the E--W approaches (straight/left).

\textbf{Duration / Action.}
At each decision step, the agent selects a phase and its duration (green time):
\[
a=(P_i,t_i),\quad P_i\in P,\; t_i\in(0,40]\ \text{s},
\]
while yellow and all-red intervals are fixed constants and not part of the action space. A detailed illustration is shown in Appendix~\ref{fig:Intersection}
%Fig.~\ref{fig:Intersection}

\section{Herald Module}
\subsection{Herald Learns Scenario-Specific Knowledge}

Herald augments LLM-based traffic signal control by learning scenario-specific dynamics from simulation. Herald adjusts phase durations as a function of queue length and logs key metrics to support short-horizon traffic prediction.

% \begin{figure}[h]
% \centering
% \includegraphics[width=\columnwidth]{figure/Figure_3.png}
% \caption{Learning architecture of Herald Module}
% \label{fig:HeraldLearn}
% \end{figure}
For each approach, Herald records maximum speed, mean egress time, and the release time associated with a given queue length (the interval from the departure of the first vehicle to the moment the last vehicle is entirely outside the incoming lane). Measuring release time over a range of queue lengths yields a practical mapping from queue length to release time while abstracting vehicle-level details (e.g., length, acceleration).

Based on the collected data, a monotonic mapping from queue length to release time is estimated and represented with a piecewise-linear model to balance fidelity and parsimony. The mapping supports second-level, fine grained signal control and guides the LLMs in selecting phase durations under varying traffic conditions.

\subsection{Herald Module Assist Process for Duration Selection}

Herald Module determines signal phase durations by combining a queue-based release time model with predicted arrivals of running vehicles.

At each timestep, for every phase \(P_i\) (\(i=1,\dots,4\)), a reference release time \(t_{\text{ref},P_i}\) is obtained from the learned mapping from queue length to release time under the current queue state, providing the baseline duration for phase \(P_i\).

\begin{figure}[t]
\centering
\includegraphics[width=\columnwidth]{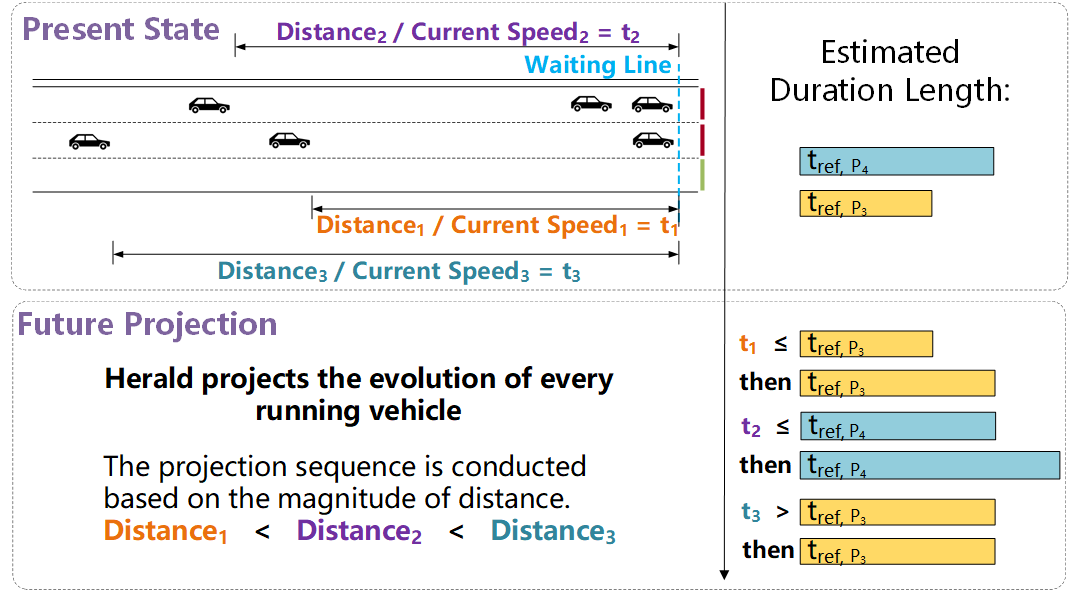}
\caption{Herald Module Assist for Duration Calculation}
\label{fig:HeraldCal}
\end{figure}

Running vehicles approaching the intersection are then evaluated. For each vehicle \(v\) associated with phase \(P_i\), the time-to-waiting-line \(\tau_v\) is estimated from current distance and speed. A vehicle \(v\) is admitted to phase \(P_i\) if \(\tau_v \le t_{\text{ref},P_i}\). Equivalently,
\[
\mathcal{A}_i=\{\, v \in \mathcal{V}_i : \tau_v \le t_{\text{ref},P_i} \,\}.
\]
Phase \(P_i\) is extended to serve vehicles in \(\mathcal{A}_i\) (admitted set); otherwise, the baseline duration is retained. As illustrated in Fig.~\ref{fig:HeraldCal}

By accounting for both queued and approaching vehicles, Herald supplies assistive, online phase duration estimates to distributed LLM-based agents as conditions evolve.

\section{HeraldLight Framework}

The phase selection prompt is a structured template composed of six blocks:
\(d_{\text{intersection}}\) (static geometry and lane--phase mapping),
\(d_{\text{task}}\) (objectives: phase choice and duration setting),
\(d_{\text{details}}\) (computational rules),
\(d_{\text{req}}\) (I/O constraints and formatting),
\(d_{\text{rules}}\) (fallbacks for edge cases),
and \(d_{\text{important}}\) (priority constraints to enforce consistency).
% A complete template is given in Appendix~\ref{tab:prompt-template-merged}
%Table~\ref{tab:prompt-template-merged}

\subsection{Task Description (\(d_{\text{task}}\))}

Two objectives are addressed: (i) select the next signal phase and (ii) determine its duration. Herald provides the quantities needed by the LLMs to reason over current and predicted demand.

\subsubsection{Signal Phase Selection}

Let \(P_i\) (\(i=1,\ldots,4\)) denote the phases and \(Q_{P_i,1},Q_{P_i,2}\) the lane queues served by \(P_i\).
The total demand for phase \(i\) is
\begin{align}
Q_i = Q_{P_i,1}+Q_{P_i,2},
\end{align}
with \(Q_i=\max(Q_{P_i,1},Q_{P_i,2})\) when one lane is empty.
Lane imbalance is quantified by
\begin{align}
I_i = |\,Q_{P_i,1}-Q_{P_i,2}\,|.
\end{align}
Phase priority is computed from \((Q_i,I_i)\).
Two queue sources are available: a Herald prediction (forward looking) and an Original measurement (current state). The predictive source is used by default; the original measurement is invoked by \(d_{\text{rules}}\) when prediction implies implausible durations or excessive imbalance.

\subsubsection{Duration Determination}

Herald supplies a mapping from queue length to release time, yielding a reference duration for each phase.
For phase \(P_i\), the reference duration is
\begin{align}
t_{\mathrm{ref}} = A\,\tau + \delta, \qquad
A=\max\!\bigl(Q_{P_i,1},\,Q_{P_i,2}\bigr),
\end{align}
where \(\tau\) is the mean per-vehicle egress time and \(\delta\) corrects for tail speed-up effects.
Then piecewise adjustments are conducted to refine the reference value:
\begin{align}
t_{\mathrm{ref}} =
\begin{cases}
t_{\mathrm{ref}} + \kappa_{\mathrm{over}}, & \text{if } t_{\mathrm{ref}} \ge T_{\mathrm{over}},\\
t_{\mathrm{ref}} + \kappa_{\mathrm{ineff}}, & \text{if } t_{\mathrm{ref}} = T_{\mathrm{ineff}},\\
\vdots\\
t_{\mathrm{ref}}, & \text{otherwise}.
\end{cases}
\end{align}
Thresholds \(T_{\mathrm{over}}, T_{\mathrm{ineff}}\) and gains \(\kappa_{\mathrm{over}}, \kappa_{\mathrm{ineff}}\) are provided by Herald Module.
Durations are computed under both predictive (\emph{Herald}) and measured (\emph{Original}) queues, and the final actions are selected by LLM-based distributed agents consistently across phases to balance throughput and fairness.

% , formally defined as,
% \begin{flalign}
% \min_{\pi_1, \pi_2, ..., \pi_I} \quad
% \lambda_1 \cdot \bar{QL} + \lambda_2 \cdot \bar{TT} + \lambda_3 \cdot \bar{WT} + \lambda_4 \cdot \bar{ET}, \label{eq_1}
% \end{flalign}
% \emph{in which,}
% \begin{flalign}
% \bar{QL} &= \frac{1}{T \times  I  \times UL_i} \sum_{t=1}^T \sum_{i=1}^I \sum_{j=1}^{UL_i} QL(ul_i^j \mid t) \tag{\ref{eq_1}{a}} \label{eq_1a} \\
% \bar{TT} &= \frac{1}{V} \sum_{v=1}^V TT_v \tag{\ref{eq_1}{b}} \label{eq_1b} \\
% \bar{WT} &= \frac{1}{V} \sum_{v=1}^V WT_v \tag{\ref{eq_1}{c}} \label{eq_1c} \\
% \bar{ET} &= \frac{1}{E} \sum_{e=1}^E ET_e \tag{\ref{eq_1}{d}} \label{eq_1d}
% \end{flalign}

\subsection{Distributed Dual LLMs Agents}
In multi-intersection settings, a distributed, parallel inference architecture deploys a population of agents. Task-specific agents are constructed via LoRA-based fine-tuning combined with dual LLMs error-corrective learning, and execute parallel inference at the action decision time of the dynamic signal-timing loop for each intersection, enabling low-latency, consistent control across all scenarios\cite{10763793}.

\subsubsection{Imitation Fine-tuning with LoRA}

An online LLM (ChatGPT) interacts with CityFlow through Herald-assisted prompts. Herald Module encodes the evolving simulator state into compact, model-aligned text. The LLM outputs actions in a schema (e.g., \mbox{\texttt{<signal>ETWT</signal>}}\allowbreak\mbox{\texttt{<duration>15</duration>}}); a lightweight parser extracts the phase and duration and executes them in CityFlow, enabling closed-loop control.

Each step logs the structured prompt, model response, and executed action to build an imitation corpus. Low-Rank Adaptation (LoRA)~\cite{hu2022lora} inserts low-rank adapters into selected layers and updates parameters, reducing computation and storage while preserving the base model. Applied to Llama-3.1-8B, LoRA yields a task-specialized LLM-Agent for traffic-signal control with modest resource demands and improved inference efficiency.

\subsubsection{Dual LLMs architecture}

\begin{figure}[t]
\centering
\includegraphics[width=\columnwidth]{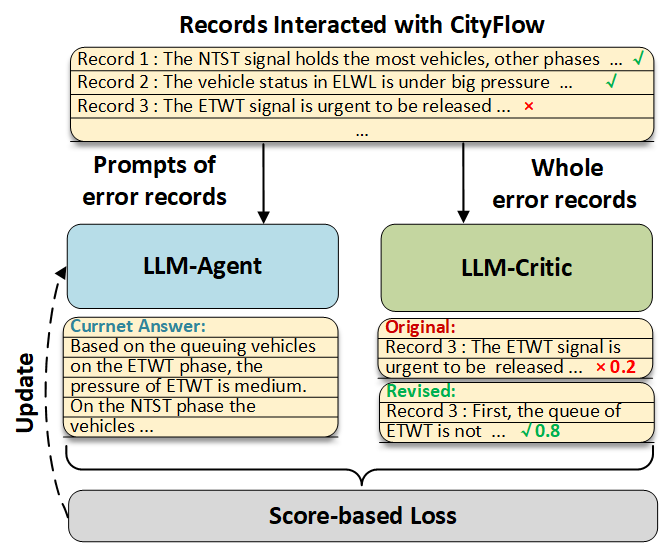}
\caption{Score-Based Dual LLMs Correction}
\label{fig:DualLLM}
\end{figure}

To mitigate hallucination in large models~\cite{10.1145/3703155}, a dual LLMs architecture is adopted: LLM-Agent generates actions; LLM-Critic (a stronger reasoner, instantiated with the ChatGPT) evaluates outputs, proposes corrections, and assigns quality scores (shown in Fig.~\ref{fig:DualLLM}). Interaction logs from CityFlow provide prompts, agent responses, and outcomes. The critic identifies erroneous cases, produces corrected responses, and pairs \{original, corrected\} with scores to create preference-style training data.

Let \(Y_i\) denote a revised (critic-corrected) reasoning trajectory for input \(X\), and \(\pi_\theta\) the agent policy. The average token log-likelihood is
\begin{align}
p_i \;=\; \frac{1}{|Y_i|}\sum_{w \in Y_i} \log \pi_\theta\bigl(y_{i,w}\mid X, Y_i{<}w\bigr).
\end{align}
A score-based ranking loss~\cite{wang2024making} prioritizes higher-scored trajectories over lower-scored ones:
\begin{align}
L_{\text{score}} \;=\; \log\!\Bigg(1 + \sum_{q_i > q_j} \Big[ e^{(p_j - p_i)} \;+\; e^{\big(2 p_{j^*} - 2\beta - p_i - p_j\big)} \Big] \Bigg),
\end{align}
where \(q_i\) is the critic-assigned quality ($Q_{\text{error/corrected}}$) for \(Y_i\), \(p_{j^*} = \min_{q_k > q_j} p_k\) is the least favorable higher-rated path, and \(\beta\) is a margin hyperparameter. The first term increases preference for higher-rated trajectories; the second term prevents performance degradation.

Training alternates among interaction data collection, critic-guided correction, and updates of the LLM-Agent using \(L_{\text{score}}\). Leveraging erroneous and corrected actions aligns the agent with critic-calibrated behavior, reducing hallucinations and improving decision reliability. The overall procedure is shown in Algorithm~\ref{alg:dual_llm}. This regimen improves robustness when the LLM-Agent performs parallel inference across multiple intersections with asynchronous decision steps.

\begin{algorithm}[h]
\caption{Score-Based Dual LLMs Decision and Fine-tuning Process}
\label{alg:dual_llm}
\KwIn{State $s(t)$, Pre-trained $\text{LLM}_{\text{Agent}}$, $\text{LLM}_{\text{Critic}}$}
\KwOut{Updated $\text{LLM}_{\text{Agent}}$}

\textbf{Initialize:} Simulation environment and $\text{LLM}_{\text{Agent}}$\;

\For{\textnormal{each episode}}{
  $s(t)\gets\text{Environment Initialization}$; $\text{Prompt}\gets\text{Conversion}(s(t))$\;

  \For{$t=1$ \KwTo $T$}{
    $a_1(t)\gets\text{LLM}_{\text{Agent}}(\text{Prompt})$; $\text{Execute}(a_1(t))\leftarrow s(t+1)$; $\text{Record}(\text{Prompt},a_1(t))$; \lIf{$a_1(t)$ fails}{$\text{Mark}(a_1(t), \text{Hallucinated})$}
  }

\For{each hallucinated $a_1(t)$}{
  $a_2(t)\gets\text{LLM}_{\text{Critic}}(\text{Prompt},a_1(t))$\;
  $\text{Add to Fine-tuning Dataset}$\\
  \Indp $((\text{Prompt},a_1(t)),a_2(t))$\;\Indm
}

  \For{each $(a_1(t),a_2(t))$}{
    $q_1\gets Q_{error}$; $q_2\gets Q_{corrected}$\;
    $L_{\text{score}}\gets\log\!\Big(1+\sum_{q_i>q_j}\big[e^{(p_j-p_i)}+e^{(2p_{j*}-2\beta-p_i-p_j)}\big]\Big)$\;
    $\text{Finetune}(\text{LLM}_{\text{Agent}})$\;
  }
}
\end{algorithm}

\section{Experiments and Results}%3 

\label{subsec1}

The evaluation of HeraldLight is organized around three research questions:\\
\noindent\textbf{RQ1:} Effectiveness relative to a state-of-the-art method.\\
\noindent\textbf{RQ2:} Generalization across cities and traffic densities (scalability and transferability).\\
\noindent\textbf{RQ3:} Robustness to hallucinations and interpretability.

\subsection{Datasets and Scenario Configuration}

Benchmarks use CityFlow datasets~\cite{wei2019survey} from three regions (datasets: https://traffic-signal-control.github.io/\#open-datasets).

\begin{figure}[h]
  \centering
  \begin{minipage}[t]{0.32\linewidth}
    \centering\includegraphics[width=\linewidth]{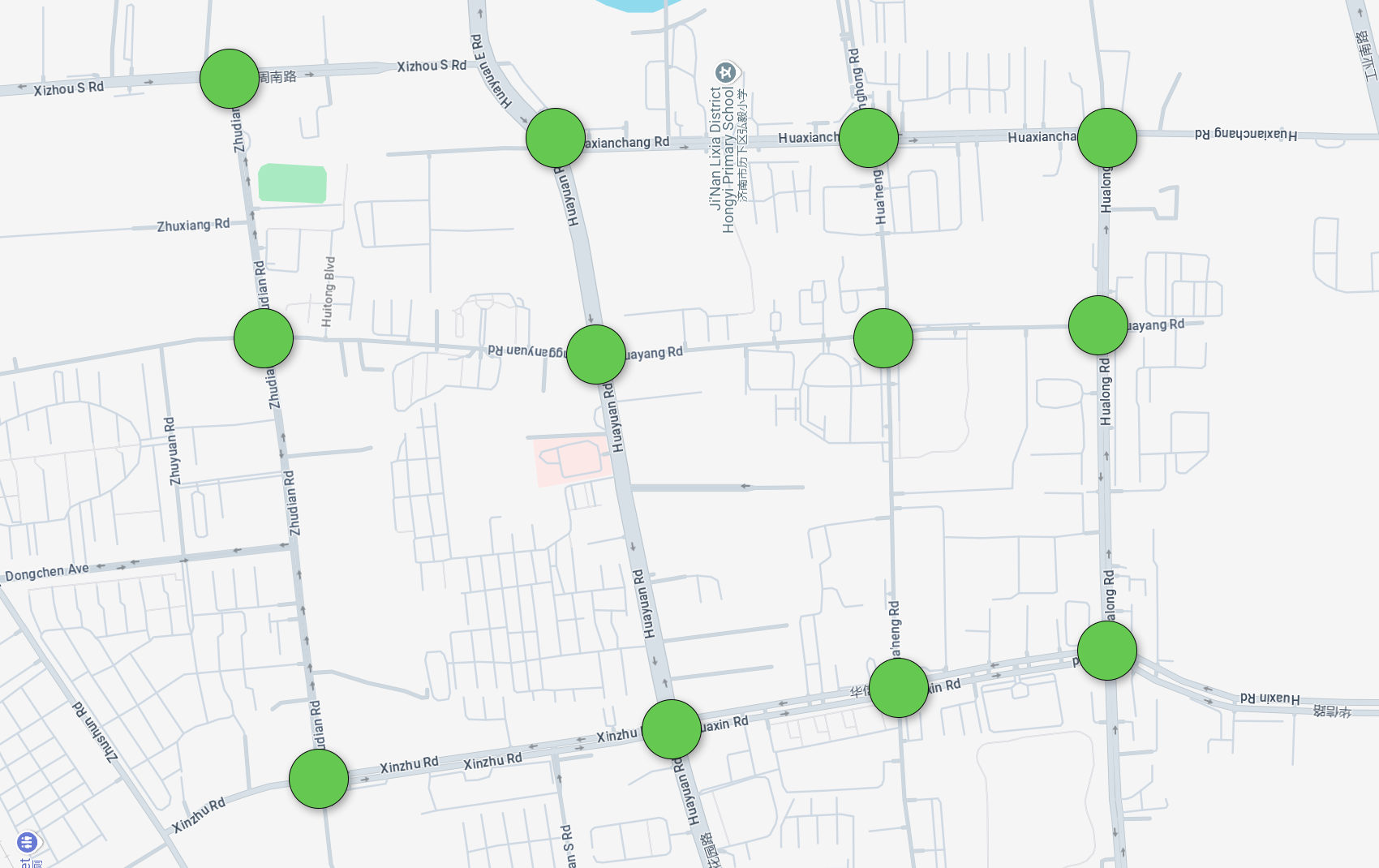}\\
    \scriptsize (a) Dongfeng (Jinan)
  \end{minipage}\hfill
  \begin{minipage}[t]{0.32\linewidth}
    \centering\includegraphics[width=\linewidth]{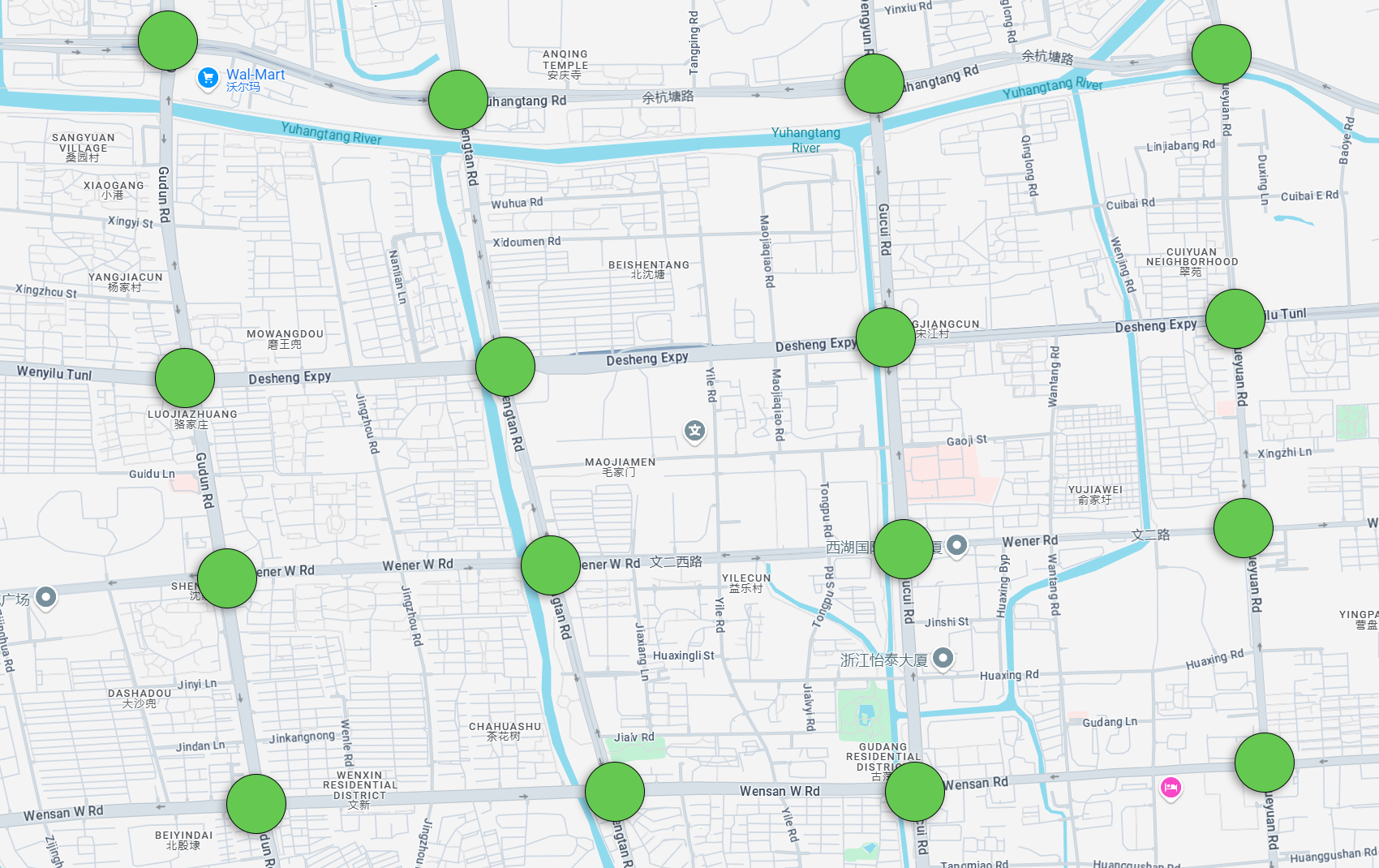}\\
    \scriptsize (b) Gudang (Hangzhou)
  \end{minipage}\hfill
  \begin{minipage}[t]{0.32\linewidth}
    \centering\includegraphics[width=\linewidth]{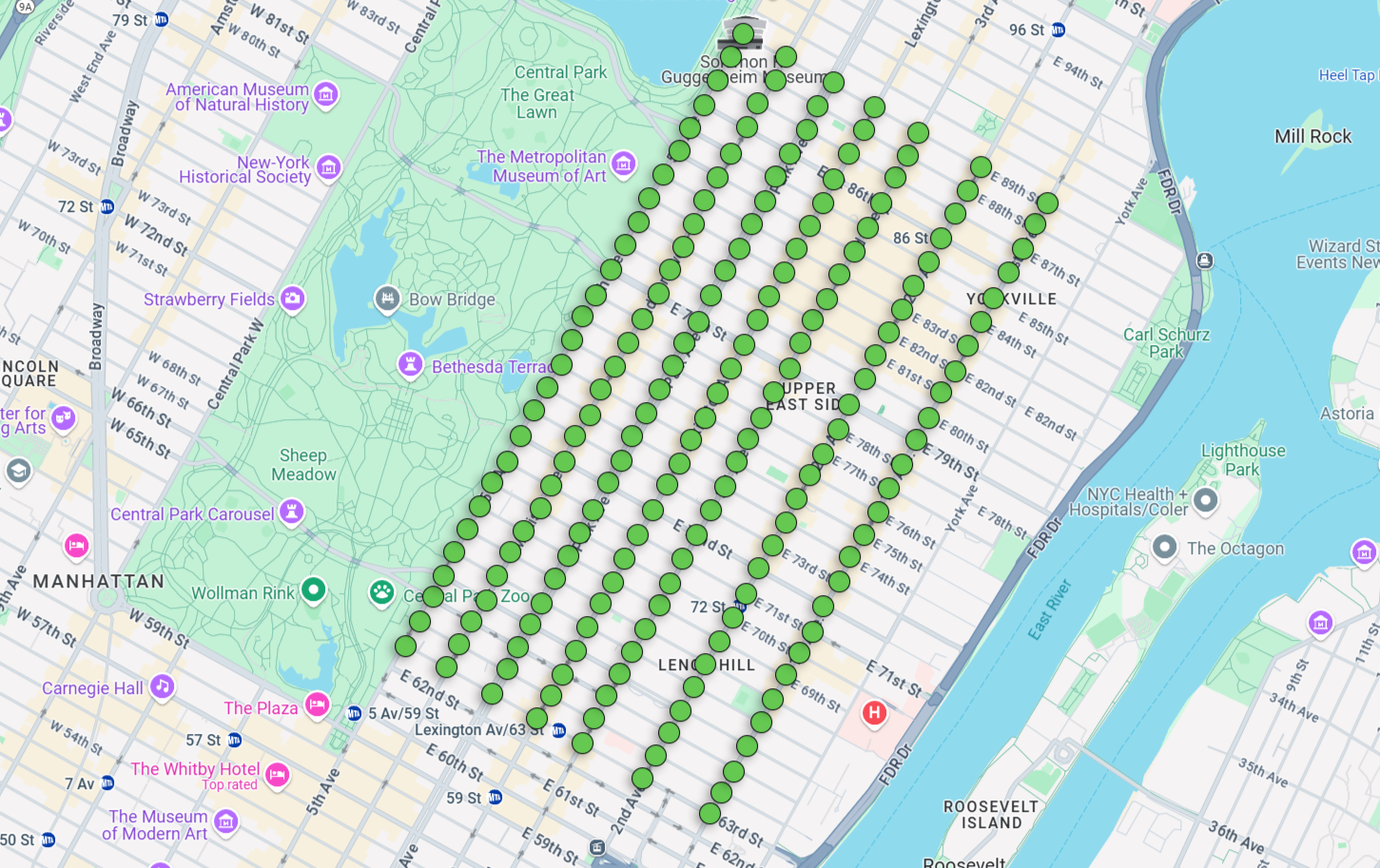}\\
    \scriptsize (c) Upper East Side (New York)
  \end{minipage}
  \caption{Visualizations of road networks in three study areas.}
  \label{fig:RoadNetworks}
\end{figure}

\noindent\textbf{Jinan} (Dongfeng; Fig.~\ref{fig:RoadNetworks}(a)): 12 intersections; vehicles: \(6295/4365/5494\).\\
\noindent\textbf{Hangzhou} (Gudang; Fig.~\ref{fig:RoadNetworks}(b)): 16 intersections; vehicles: \(2983/6984\).\\
\noindent\textbf{New York} (Upper East Side; taxi-trip derived; Fig.~\ref{fig:RoadNetworks}(c)): 196 intersections; vehicles: \(11058/16337\).

% Add \usepackage{graphicx} to your preamble if you don't have it already
\begin{table*}[t]
\centering
\caption{Results on \textbf{Jinan} (JN-1/2/3) and \textbf{Hangzhou} (HZ-1/2) across three metrics. All values are rounded to two decimal places. The best results are in \textbf{bold}, the second-best are \underline{underlined}, and the third-best are \underline{\underline{double underlined}}.}
\label{tab:comparison-all-rounded}
\begingroup
%\fontsize{10pt}{12pt}\selectfont 
\setlength{\tabcolsep}{2.5pt}      % Adjust column spacing
\renewcommand{\arraystretch}{0.94} % Adjust row spacing
\begin{tabular}{@{}l*{5}{ccc}@{}}
\toprule
Method & \multicolumn{3}{c}{JN-1} & \multicolumn{3}{c}{JN-2} & \multicolumn{3}{c}{JN-3} & \multicolumn{3}{c}{HZ-1} & \multicolumn{3}{c}{HZ-2} \\
\cmidrule(r){2-4}\cmidrule(lr){5-7}\cmidrule(lr){8-10}\cmidrule(lr){11-13}\cmidrule(l){14-16}
& ATT & AQL & AWT & ATT & AQL & AWT & ATT & AQL & AWT & ATT & AQL & AWT & ATT & AQL & AWT \\
\midrule
\multicolumn{16}{c}{\textbf{Transportation Methods}} \\
\midrule
Random           & 584.08 & 657.86 & 97.16  & 541.95 & 405.17 & 90.44  & 548.53 & 516.38 & 101.37 & 605.63 & 287.78 & 93.52  & 524.12 & 457.20 & 109.40 \\
FixedTime        & 481.79 & 491.03 & 70.99  & 441.19 & 294.14 & 66.72  & 450.11 & 394.34 & 69.19  & 616.02 & 301.33 & 73.99  & 486.69 & 425.12 & 72.80  \\
MaxPressure      & 282.58 & 170.71 & 44.53  & 273.20 & 106.58 & \underline{\underline{38.25}} & 265.75 & 133.90 & 40.20 & 325.33 & 68.99  & 49.60  & 347.74 & 215.53 & 70.58  \\
\midrule
\multicolumn{16}{c}{\textbf{RL Methods}} \\
\midrule
MPLight          & 299.53 & 202.29 & 92.34  & 301.40 & 138.34 & 94.73  & 284.60 & 161.79 & 84.02  & 343.81 & 83.34  & 93.88  & 368.36 & 255.73 & 112.41 \\
AttendLight      & 296.97 & 195.48 & 67.72  & 281.29 & 115.36 & 56.63  & 272.56 & 143.46 & 56.11  & 324.43 & 67.41  & 58.78  & 351.12 & 223.73 & 67.89  \\
PressLight       & 288.91 & 182.28 & 51.47  & 276.70 & 111.61 & 47.06  & 274.58 & 145.45 & 41.71  & 347.39 & 85.11  & 84.45  & 395.84 & 297.85 & 129.80 \\
CoLight          & 284.26 & 175.84 & 61.80  & 274.68 & 108.41 & 51.68  & 265.67 & 134.11 & 50.41  & 319.92 & 64.67  & 55.84  & 339.79 & 206.65 & 81.94  \\
Efficient-CoLight& 276.89 & 163.20 & 42.60  & 268.85 & 102.35 & 39.35  & 261.85 & 129.06 & 40.44  & 312.14 & 58.26  & \underline{\underline{36.33}} & 328.65 & 181.81 & 56.27  \\
Advanced-CoLight & \underline{\underline{273.20}} & \underline{\underline{158.65}} & 47.26  & 267.56 & \underline{\underline{101.26}} & 40.81  & \underline{\underline{260.16}} & 127.04 & 42.84  & \underline{\underline{304.54}} & \underline{\underline{52.97}} & 40.67  & \underline{\underline{322.86}} & \underline{\underline{173.38}} & 69.93  \\
DynamicLight     & \underline{238.61} & \underline{83.48}  & \textbf{27.35}  & \underline{221.91} & \underline{32.46}  & \textbf{16.55}  & \underline{220.07} & \underline{50.23}  & \textbf{20.95}  & \underline{264.73} & \underline{13.02}  & \textbf{12.33}  & \underline{314.73} & \textbf{146.06} & \underline{53.99}  \\
\midrule
\multicolumn{16}{c}{\textbf{Large-Scale AI Models}} \\
\midrule
Llama-2-13B      & 401.92 & 368.41 & 123.61 & 406.44 & 254.05 & 150.78 & 396.29 & 324.13 & 148.24 & 468.79 & 180.01 & 183.17 & 425.96 & 342.27 & 143.18 \\
ChatGPT-3.5      & 315.95 & 222.40 & 53.72  & 279.83 & 112.77 & 43.62  & 288.98 & 163.13 & 49.49  & 324.55 & 68.25  & 46.16  & 340.29 & 196.38 & \underline{\underline{55.28}} \\
ChatGPT-4        & 276.96 & 164.33 & 48.81  & \underline{\underline{263.82}} & 133.04 & 47.58  & 271.78 & \underline{\underline{105.60}} & 45.12  & 314.91 & 59.93  & 53.27  & 333.57 & 191.38 & 67.02  \\
ChatGPT-4o-mini  & 275.00 & 161.06 & 44.83  & 267.01 & 101.03 & 41.33  & 261.14 & 128.73 & 42.89  & 313.22 & 58.78  & 42.10  & 332.47 & 186.75 & 61.01  \\
DeepSeek-R1-671B & 276.78 & 163.44 & 46.05  & 271.18 & 105.56 & 40.35  & 261.82 & 129.38 & 41.52  & 312.09 & 57.68  & 45.72  & 331.99 & 186.83 & 72.93  \\
ERNIE-4.0-8K     & 275.93 & 161.90 & 44.07  & 270.20 & 103.94 & 38.49  & 261.71 & 129.05 & 41.83  & 311.54 & 57.82  & 43.67  & 333.22 & 187.99 & 57.40  \\
ERNIE-Lite       & 301.51 & 202.62 & 63.10  & 292.92 & 126.50 & 61.81  & 285.13 & 159.83 & 59.98  & 328.38 & 70.02  & 67.58  & 346.60 & 215.39 & 84.63  \\
ERNIE-Speed-8K   & 313.75 & 222.46 & 87.56  & 301.87 & 137.58 & 83.14  & 287.35 & 163.82 & 76.14  & 338.21 & 78.55  & 99.49  & 352.92 & 230.70 & 99.38  \\
Hunyuan-Lite     & 1135.71& 1431.58& 950.23 & 1299.84& 1173.71& 1004.29& 1174.65& 1261.17& 980.14 & 1266.47& 712.10 & 1041.25& 810.36 & 795.93 & 872.42 \\
Qwen-Long        & 288.67 & 179.99 & 49.41  & 276.50 & 110.38 & 45.80  & 270.24 & 140.28 & 45.11  & 319.71 & 64.22  & 48.27  & 334.90 & 189.07 & \textbf{50.90}  \\
Spark-Lite       & 569.47 & 670.15 & 320.70 & 535.76 & 416.11 & 272.29 & 514.62 & 505.74 & 249.83 & 554.95 & 255.74 & 244.67 & 451.81 & 383.45 & 186.11 \\
\midrule
\multicolumn{16}{c}{\textbf{LLM-Based Methods}} \\
\midrule
LLMLight         & 277.01 & 162.56 & \underline{\underline{41.77}} & 269.07 & 103.23 & 43.33  & 261.42 & 128.95 & 42.12  & 313.72 & 59.58  & 51.36  & 336.99 & 194.86 & 70.87  \\
Traffic-R1         & 277.69 & 164.74 & 45.70 & 270.11 & 104.32 & 41.78  & 260.12 & 125.91 & \underline{\underline{40.18}}  & 311.01 & 57.57  & 40.99  & 331.68 & 185.12 & 58.95  \\
HeraldLight      & \textbf{234.64} & \textbf{73.69}  & \underline{40.55}  & \textbf{220.78} & \textbf{28.58}  & \underline{27.63}  & \textbf{217.69} & \textbf{44.15}  & \underline{29.96}  & \textbf{262.74} & \textbf{10.03}  & \underline{32.24}  & \textbf{297.85} & \underline{153.51} & 120.03 \\
\bottomrule
\end{tabular}
\endgroup
\end{table*}

\subsection{Simulation Platform and Configuration}
Simulations are conducted in CityFlow~\cite{Zhang2019CityFlowAM}. Each intersection runs four phases (NTST, NLSL, ETWT, ELWL) with right turns permitted; safety timing is 3\,s yellow + 2\,s all-red. Two strategies are evaluated: \emph{Fixed} (30\,s per phase) and \emph{Dynamic} (adaptive up to 40\,s). Each scenario runs for 1\,hour.

\subsubsection{Performance Metrics}
\textbf{ATT} (\(\mathrm{s}\)): average travel time; 
\textbf{AQL} (\(\mathrm{m}\)): average queue length (network-wide); 
\textbf{AWT} (\(\mathrm{s}\)): average waiting time at intersections.

\subsubsection{Compared Models}
Baselines include transportation methods (Random, FixedTime~\cite{Koonce2008}, MaxPressure~\cite{VARAIYA2013177}), deep RL families (PressLight~\cite{10.1145/3292500.3330949}, MPLight~\cite{Chen_Wei_Xu_Zheng_Yang_Xiong_Xu_Li_2020}, CoLight~\cite{10.1145/3357384.3357902}, Efficient-CoLight~\cite{Wu2021EfficientPI}, Advanced-CoLight~\cite{advanced_xlight}, AttendLight~\cite{NEURIPS2020_29e48b79}), Large-scale AI Models (Llama-2-13B, ChatGPT-3.5/4/4o-mini, ERNIE-4.0-8K/ERNIE-Lite/ERNIE-Speed-8K, Hunyuan-Lite, Qwen-Long, Spark-Lite, LLMLight~\cite{lai2023large}, Traffic-R1~\cite{zou2025trafficr1reinforcedllmsbring}), the dynamic timing method DynamicLight~\cite{zhang2024dynamiclighttwostagedynamictraffic}, and the proposed HeraldLight.
\subsubsection{Resource Usage}
Dual-socket Intel Xeon Platinum 8457C with a single NVIDIA L20 (48~GiB, CUDA~12.4). Under synchronous batching (avg. batch size $\approx$ 2.85; avg. input/output 960/356 tokens), we observe an average batch-level inference time of 9.09~s, a token throughput of 413.35~tokens/s, and GPU utilization of 95.96\% with peak memory usage of 44{,}280.6~MiB.

% \subsubsection{Model Settings}
% RL baselines share learning rate $1{\times}10^{-3}$, replay buffer $12{,}000$, and sample size $3{,}000$. LLM decoding uses top\_p $=1.0$ and temperature $=0.1$. For imitation fine-tuning (LightGPT, HeraldLight), LoRA rank $=8$ and learning rate $3{\times}10^{-4}$.

\subsection{Comparison Between HeraldLight and Other TSC Methods (RQ1)}
As shown Table~\ref{tab:comparison-all-rounded}, across five CityFlow scenarios Jinan (1-3), Hangzhou (1-2), HeraldLight, relative to a state-of-the-art baseline (DynamicLight), ranks first in ATT  and remains top-2 in AQL across five scenarios. In Jinan 1, AQL is lower than DynamicLight by \textbf{11.72\%}, whereas {DynamicLight attains lower AWT in multiple scenarios. Dynamic timing methods collectively outperform fixed timing baselines; fixed timing RL variants (e.g., Advanced-CoLight) are competitive in isolated cases but underperform on aggregate. 

Variation in performance stems from distinct mechanisms: RL improves as optimization reduces policy error and sharpens value estimates, aligning states and actions. Large-scale AI models (e.g., ChatGPT, DeepSeek) match Advanced-Colight without task-specific training due to strong reasoning priors and instruction following, while LLMLight remains competitive under fixed timing but is constrained by capacity and schedule rigidity. Dynamic timing resolutions DynamicLight uses 5-s multiples, whereas HeraldLight provides 1-s control, enabling finer timing that reduces delay and increases throughput.

\subsection{Ablation Study}

An ablation on Hangzhou 2 (high demand) with Llama-3.1-8B quantifies module contributions (Table~\ref{tab:ablation_study}). The base model and the base augmented with Herald failed to complete the task, indicating that traffic-aware forecasting alone is insufficient under heavy load. Imitation fine-tuning stabilized control (ATT 303.581). Incorporating the Dual LLMs architecture further reduced ATT to 297.852, supporting complementary gains from forecasting and critic-guided arbitration.

\begin{table}[!t]
  \centering
  \caption{Ablation on Hangzhou~2 with Llama-3.1-8B (lower is better).  \textemdash\ indicates the task was not completed.}
  \label{tab:ablation_study}
  \footnotesize
  \setlength{\tabcolsep}{4pt}
  \renewcommand{\arraystretch}{1.15}
  \resizebox{\linewidth}{!}{
  \begin{tabular}{c p{0.45\columnwidth} r r r}
    \hline
    \textbf{No.} & \textbf{Configuration} & \textbf{ATT} & \textbf{AQL} & \textbf{AWT} \\
    \hline
    1 & Llama-3.1-8B (base)                & \textemdash & \textemdash & \textemdash \\
    2 & No.\,1 + Herald Module             & \textemdash & \textemdash & \textemdash \\
    3 & No.\,2 + Imitation Fine-tuning     & 303.581 & 154.084 & 125.120 \\
    4 & No.\,3 + Dual LLMs Architecture   & 297.852 & 153.507 & 120.029 \\
    \hline
  \end{tabular}
  }
\end{table}

\subsection{Generalization Comparison (RQ2)}
\subsubsection{Scalability}

In the multi-intersection New York network, HeraldLight scales better than competing methods shown in Fig.~\ref{fig:scalability}. Across the Jinan, Hangzhou, and New York scenarios, HeraldLight achieves a 20.03\% reduction in average travel time relative to SOTA method. Herald-guided queue forecasting and LLM-based reasoning enable distributed agents to issue context-aware phase releases under dynamic timing, accelerating discharge on high-pressure approaches and reducing delay and secondary queues. The advantage is amplified on the 196 intersections New York network approximately twelvefold larger than Jinan and Hangzhou.

%ATT scalability results are shown in Fig.~\ref{fig:scalability}.
\begin{figure}[H]
\centering
\includegraphics[width=\columnwidth]{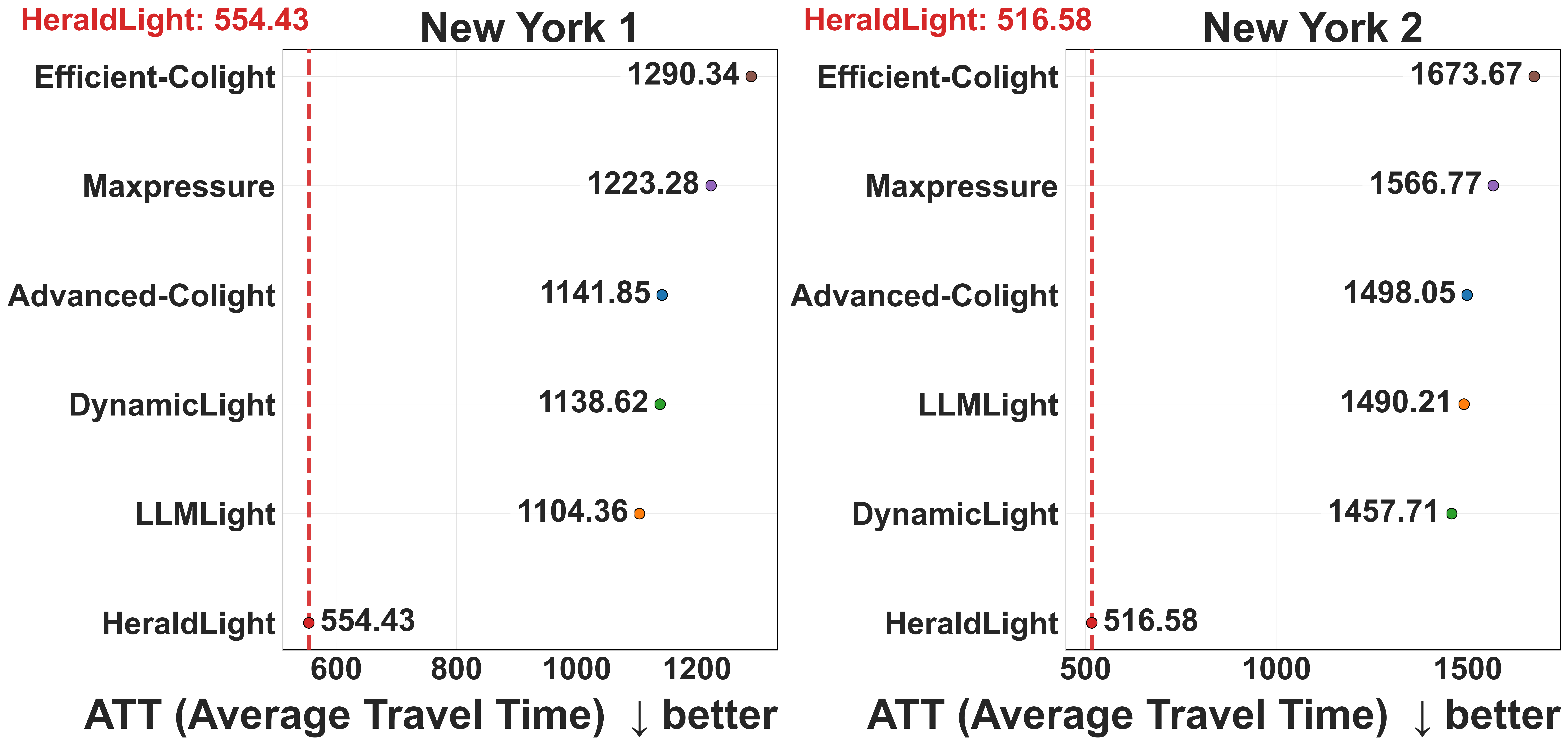}
\caption{The performance in the large-scale road network}
\label{fig:scalability}
\end{figure}

\subsubsection{Transferability}
\begin{figure}[h]
\centering
\includegraphics[width=\columnwidth]{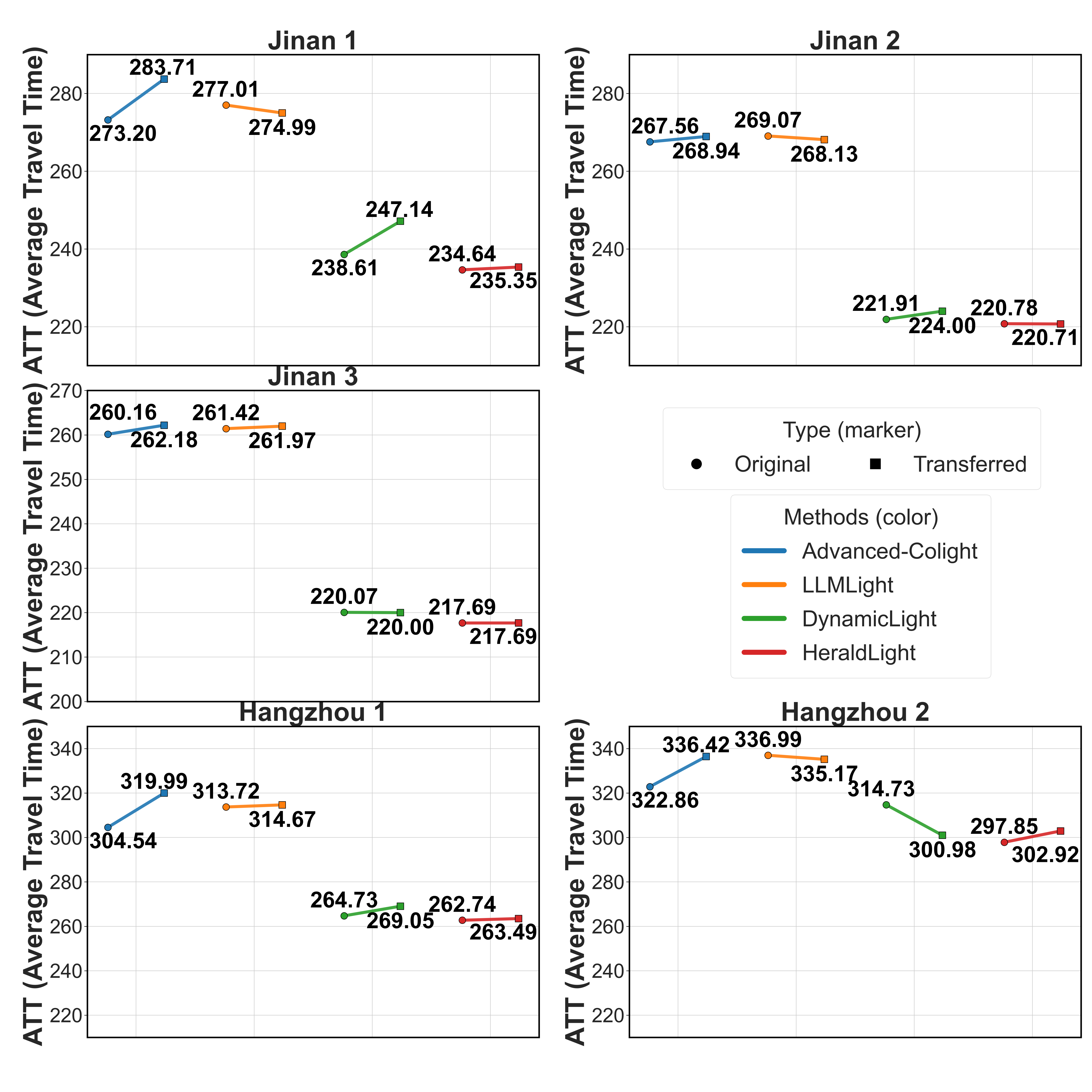}
\caption{Transferability compared in Jinan and Hangzhou}
\label{fig:Transferability}
\end{figure}

As shown in Fig.~\ref{fig:Transferability}, The original model is trained and evaluated in the target scenario, whereas the transferred model uses the same method trained on other scenarios and is evaluated in the target scenario.

Across Jinan (1-3) and Hangzhou (1-2), in-scenario training generally yields better performance than cross-scenario transfer. HeraldLight exhibits the strongest transferability: ATT differences between in-scenario and transferred models are typically below one second; in Jinan 3 the difference is 0.006 s (217.68729 and 217.69368). LLMLight ranks second, with most gaps in the range of 2–3 s and stable variability. DynamicLight and Advanced-CoLight transfer less effectively than the LLM-based agents; DynamicLight ranks third and approaches HeraldLight in Jinan 3. The marked separation between LLMLight and DynamicLight is consistent with their fixed timing and dynamic-timing designs.

\subsection{Reducing Hallucination (RQ3)}
Hallucination is defined as context-incoherent, repetitive, or constraint-violating outputs at the decision step. The metric is computed as the fraction of interactions that exhibit any such symptoms. Hangzhou~2 is the most challenging setting in training due to higher demand and more intersections.

\begin{table}[h]
\centering
\caption{Hallucination comparison on Hangzhou~2.}
\label{tab:hallucination_comparison}
\footnotesize
\setlength{\tabcolsep}{5pt}
\renewcommand{\arraystretch}{1.15}
\begin{tabular}{lccc}
\hline
\textbf{Method} & \textbf{Interactions} & \textbf{Hallucinations} & \textbf{Rate (\%)} \\ \hline
LLM-Agent   & 6048 & 558 & 9.23 \\
HeraldLight & 5525 & 9   & 0.163 \\ \hline
\end{tabular}
\end{table}

Results indicate a substantial reduction with HeraldLight: hallucination cases drop from 558 to 9 (98.4\% relative decrease) and the rate falls from 9.23\% to 0.163\%. The interaction count also declines (6048~$\rightarrow$~5525, $\sim$8.6\%), suggesting more stable decision making. These outcomes attribute the improvement to the Dual LLMs architecture and traffic-aware guidance, which increase interpretability and enhance reliability under heavy traffic.

}

\subsection{Extreme Weather Condition}

Based on the performance presented in Table~\ref{tab:comparison-all-rounded}, 
the top three methods, HeraldLight, DynamicLight, and Advanced-Colight, 
are selected according to performance under normal weather conditions. 
Stability under extreme weather conditions was then assessed by adjusting vehicle parameters 
to simulate the effects of adverse weather. 
Key modifications included a 50\% reduction in maximum positive acceleration, 
a 30\% reduction in negative acceleration, and a 30\% reduction in maximum speed.

Under extreme conditions, HeraldLight shows average increases of 41.61 s\ in ATT, 63.39\% in AQL, and 41.946\% in AWT. DynamicLight experienced greater degradation (42.83\% ATT, 81.75\% AQL, 94.23\% AWT), while Advanced-Colight showed smaller increases (34.28\% ATT, 15.67\% AQL, 30.47\% AWT). HeraldLight's transparent decision-making process allows it to quickly adjust parameters like vehicle speed and queue maps, making it more adaptable to extreme weather. 

\begin{table}[t]
\centering
\caption{EW (Extreme Weather) vs. Base per scene/metric. Each cell shows Base $\rightarrow$ EW and the percentage change vs. its own Base. Bold indicates the best results (lower is better) within each row.}
\label{tab:ew-delta-compact-process}

% --- spacing controls (no extra packages) ---
{
 \setlength{\tabcolsep}{2.5pt}           % horizontal padding per column
% 间距变量：改这里即可，如 2pt/3pt/0.6ex 等
\newcommand{\HLgap}{2pt}
% 带上下留白的横线
\newcommand{\myhline}{\noalign{\vskip\HLgap}\hline\noalign{\vskip\HLgap}}
\begin{tabular}{@{}llccc@{}}
\myhline
Scene & Metric & HeraldLight & Advanced-CoLight & DynamicLight \\
\myhline
JN-1 & ATT & \shortstack{234.64 $\rightarrow$ \textbf{361.18}\\ ($\Delta$ +53.93\%)}
           & \shortstack{273.20 $\rightarrow$ 390.18\\ ($\Delta$ +42.82\%)}
           & \shortstack{238.61 $\rightarrow$ 372.36\\ ($\Delta$ +56.05\%)} \\
JN-1 & AQL & \shortstack{73.69 $\rightarrow$ \textbf{161.51}\\ ($\Delta$ +119.17\%)}
           & \shortstack{158.65 $\rightarrow$ 218.21\\ ($\Delta$ +37.54\%)}
           & \shortstack{83.48 $\rightarrow$ 188.37\\ ($\Delta$ +125.65\%)} \\
JN-1 & AWT & \shortstack{40.55 $\rightarrow$ 84.05\\ ($\Delta$ +107.27\%)}
           & \shortstack{47.26 $\rightarrow$ \textbf{66.64}\\ ($\Delta$ +41.01\%)}
           & \shortstack{27.35 $\rightarrow$ 76.68\\ ($\Delta$ +180.37\%)} \\
\myhline
JN-2 & ATT & \shortstack{220.78 $\rightarrow$ \textbf{310.49}\\ ($\Delta$ +40.63\%)}
           & \shortstack{267.56 $\rightarrow$ 353.57\\ ($\Delta$ +32.15\%)}
           & \shortstack{221.91 $\rightarrow$ 317.60\\ ($\Delta$ +43.12\%)} \\
JN-2 & AQL & \shortstack{28.58 $\rightarrow$ \textbf{45.74}\\ ($\Delta$ +60.04\%)}
           & \shortstack{101.26 $\rightarrow$ 106.97\\ ($\Delta$ +5.64\%)}
           & \shortstack{32.46 $\rightarrow$ 57.28\\ ($\Delta$ +76.46\%)} \\
JN-2 & AWT & \shortstack{27.63 $\rightarrow$ 33.32\\ ($\Delta$ +20.59\%)}
           & \shortstack{40.81 $\rightarrow$ 51.54\\ ($\Delta$ +26.29\%)}
           & \shortstack{16.55 $\rightarrow$ \textbf{27.82}\\ ($\Delta$ +68.10\%)} \\
\myhline
JN-3 & ATT & \shortstack{217.69 $\rightarrow$ \textbf{317.98}\\ ($\Delta$ +46.07\%)}
           & \shortstack{260.16 $\rightarrow$ 351.53\\ ($\Delta$ +35.12\%)}
           & \shortstack{220.07 $\rightarrow$ 325.67\\ ($\Delta$ +47.98\%)} \\
JN-3 & AQL & \shortstack{44.15 $\rightarrow$ \textbf{84.39}\\ ($\Delta$ +91.14\%)}
           & \shortstack{127.04 $\rightarrow$ 146.94\\ ($\Delta$ +15.66\%)}
           & \shortstack{50.23 $\rightarrow$ 101.35\\ ($\Delta$ +101.77\%)} \\
JN-3 & AWT & \shortstack{29.96 $\rightarrow$ 48.99\\ ($\Delta$ +63.52\%)}
           & \shortstack{42.84 $\rightarrow$ 54.83\\ ($\Delta$ +27.99\%)}
           & \shortstack{20.95 $\rightarrow$ \textbf{40.60}\\ ($\Delta$ +93.79\%)} \\
\myhline
HZ-1 & ATT & \shortstack{262.74 $\rightarrow$ \textbf{362.75}\\ ($\Delta$ +38.06\%)}
           & \shortstack{304.54 $\rightarrow$ 399.88\\ ($\Delta$ +31.31\%)}
           & \shortstack{264.73 $\rightarrow$ 372.42\\ ($\Delta$ +40.68\%)} \\
HZ-1 & AQL & \shortstack{10.03 $\rightarrow$ \textbf{15.37}\\ ($\Delta$ +53.24\%)}
           & \shortstack{52.97 $\rightarrow$ 56.92\\ ($\Delta$ +7.46\%)}
           & \shortstack{13.02 $\rightarrow$ 26.72\\ ($\Delta$ +105.22\%)} \\
HZ-1 & AWT & \shortstack{32.24 $\rightarrow$ 33.65\\ ($\Delta$ +4.37\%)}
           & \shortstack{40.67 $\rightarrow$ 57.61\\ ($\Delta$ +41.65\%)}
           & \shortstack{12.33 $\rightarrow$ \textbf{26.98}\\ ($\Delta$ +118.82\%)} \\
\myhline
HZ-2 & ATT & \shortstack{297.85 $\rightarrow$ \textbf{385.30}\\ ($\Delta$ +29.36\%)}
           & \shortstack{322.86 $\rightarrow$ 419.83\\ ($\Delta$ +30.03\%)}
           & \shortstack{314.73 $\rightarrow$ 397.51\\ ($\Delta$ +26.30\%)} \\
HZ-2 & AQL & \shortstack{153.51 $\rightarrow$ \textbf{143.17}\\ ($\Delta$ -6.74\%)}
           & \shortstack{173.38 $\rightarrow$ 194.29\\ ($\Delta$ +12.06\%)}
           & \shortstack{146.06 $\rightarrow$ 145.53\\ ($\Delta$ -0.36\%)} \\
HZ-2 & AWT & \shortstack{120.03 $\rightarrow$ 136.81\\ ($\Delta$ +13.98\%)}
           & \shortstack{69.93 $\rightarrow$ 80.72\\ ($\Delta$ +15.43\%)}
           & \shortstack{53.99 $\rightarrow$ \textbf{59.37}\\ ($\Delta$ +9.96\%)} \\
\myhline
\end{tabular}
}% end local spacing group
\end{table}

\subsection{Actions Analysis of HeraldLgiht and DynamicLight}

% 文中
\begin{figure}[t]
  \centering
  % (a)
  \subfloat[Phase distribution]{%
    \includegraphics[width=\columnwidth]{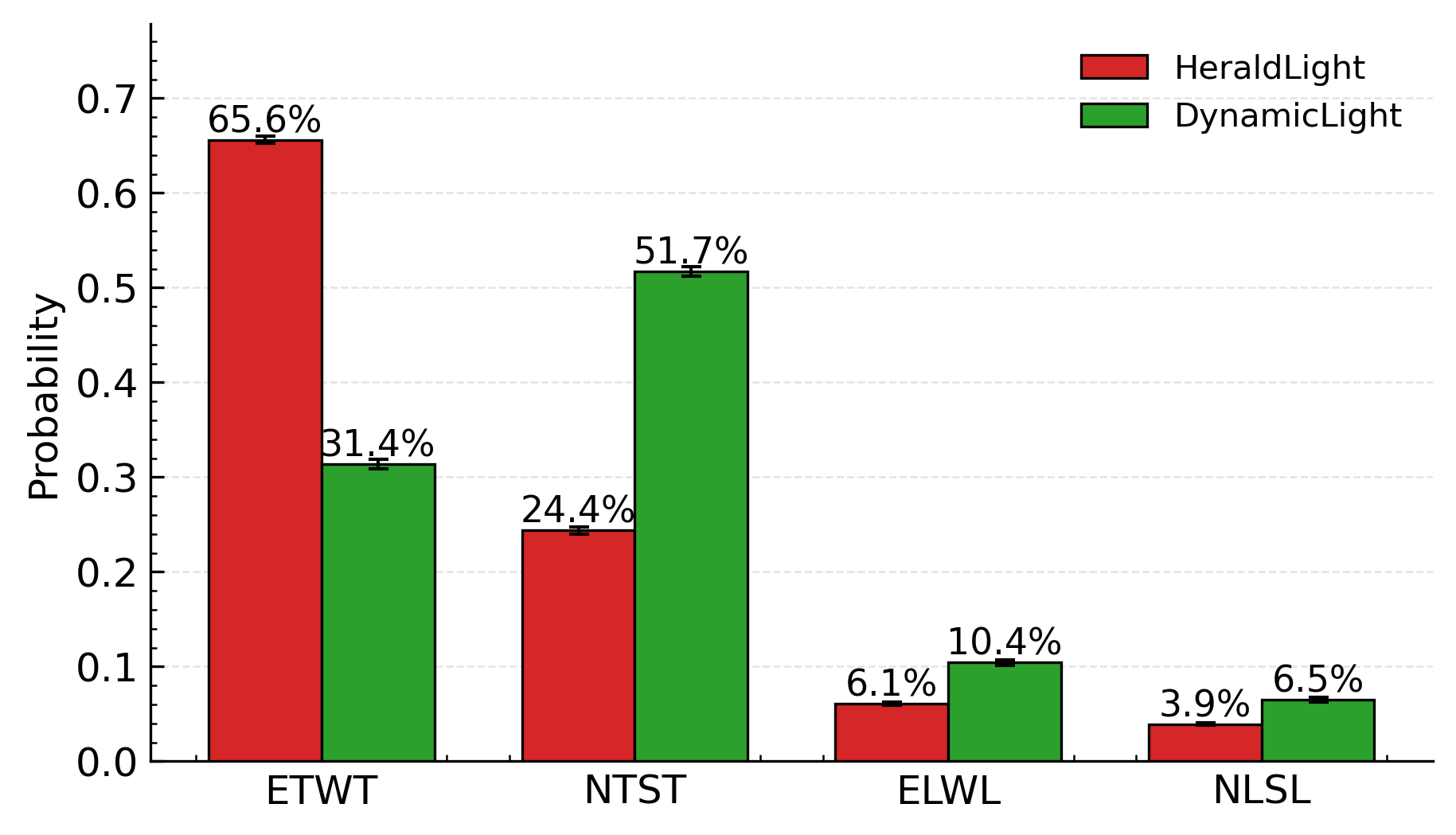}%
    \label{fig:phase_ecdf_a}%
  }\\[-0.2em] % 调整两子图间距（可微调或删去）
  % (b)
  \subfloat[Duration ECDF]{%
    \includegraphics[width=\columnwidth]{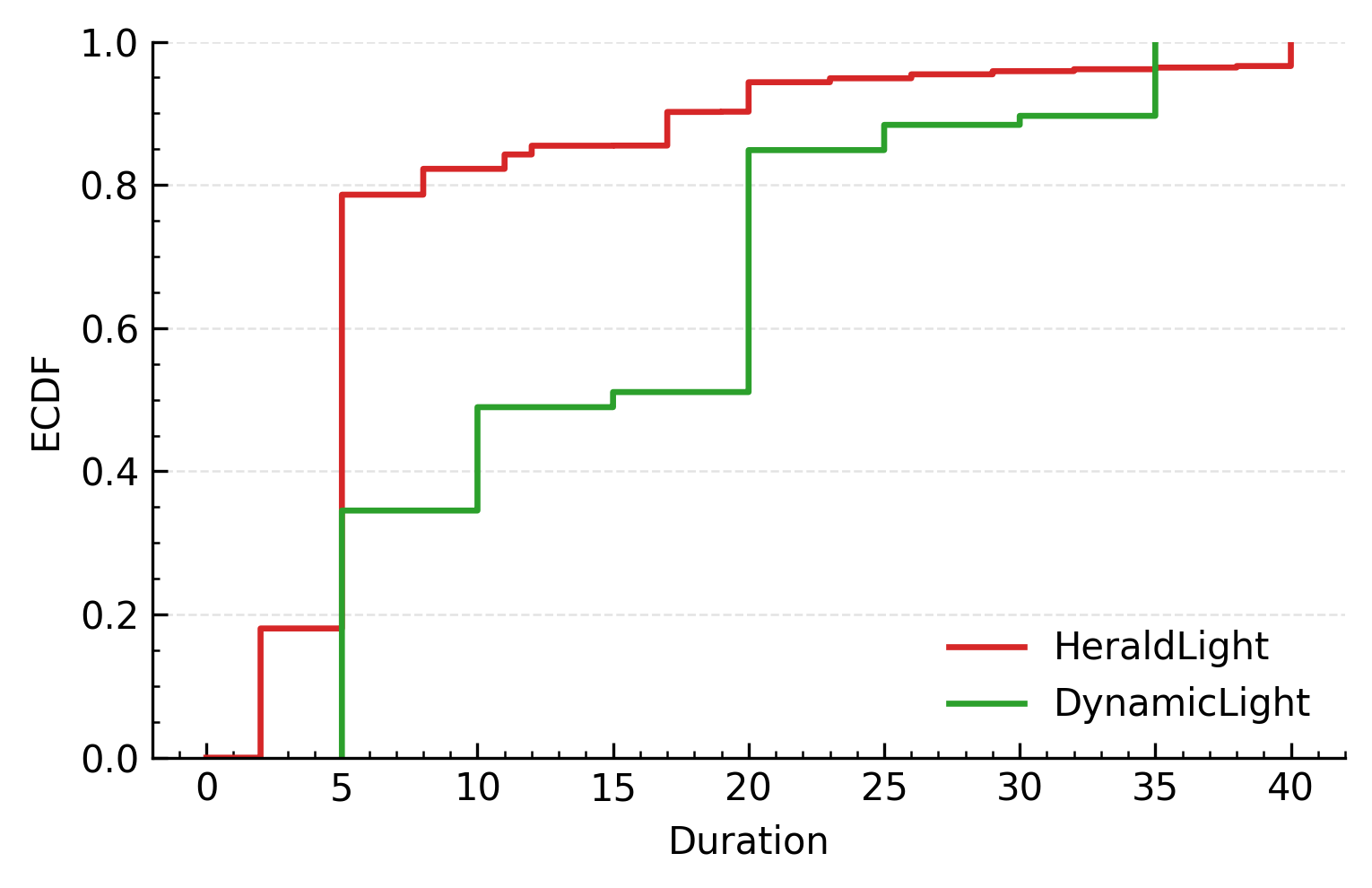}%
    \label{fig:phase_ecdf_b}%
  }
  \caption{HeraldLight vs DynamicLight action comparison (a) Phase distribution. (b) Duration ECDF (Empirical Cumulative Distribution Function).}
  \label{fig:phase_ecdf_combo}
\end{figure}

% --- Concise write-up for panels (a) and (b) ---
\noindent\textbf{Action selection in New York 1:}
In the New York grid (\(7\times28\), \(196\) intersections), we compare HeraldLight and DynamicLight. 
From Fig.~\ref{fig:phase_ecdf_combo}\subref{fig:phase_ecdf_a}, both methods emphasize ETWT and NTST phases. 
HeraldLight favors ETWT phase releases, aligning with heavier inflows along the 28 intersections east--west spine; high-volume arterials benefit from faster phase switching to sustain progression. 
Its finer control granularity enables more precise actions during peaks. 
DynamicLight biases NTST phase releases, which can aid longer cross-network trips, but during peaks this intensifies queues on east--west approaches, increasing average travel time; HeraldLight is therefore more efficient under congestion.

\medskip
\noindent\textbf{Duration distributions.}
From the ECDF in Fig.~\ref{fig:phase_ecdf_combo}\subref{fig:phase_ecdf_b}, HeraldLight's curve is smoother, indicating finer timing resolution, whereas DynamicLight exhibits 5\,s step increments (coarser granularity). 
HeraldLight concentrates green durations below \(20\,\mathrm{s}\), improving responsiveness to rapid peak fluctuations. 
DynamicLight allocates longer (mostly \(\leq 35\,\mathrm{s}\)), which elevates the ATT of the other three phases during peaks, consistent with its weaker performance in New York 1.

\section{Conclusion}
This study presents HeraldLight, a traffic signal control framework that combines a scene-aware Herald Module with large language model agents to realize dynamic, second-level timing. Findings indicate that: (i) fixed time strategies lack the precision required for fine grained control in complex traffic; (ii) coupling LLM reasoning with the Herald Module enables precise timing and improves control effectiveness; and (iii) LLMs are susceptible to hallucination, which is effectively mitigated by the proposed Dual LLMs architecture, enhancing stability and reliability. Future work will emphasize online self-evolution within the LLMs for action reasoning and output generation, and target deployment in real-world systems.

\section*{Acknowledgment}%1

This work was supported by the National Natural Science Foundation of China (Grant No. 62176024).

\appendices
% \renewcommand{\thefigure}{A.\arabic{figure}}
% \renewcommand{\thetable}{A.\arabic{table}}

% 改 caption 名称，区分 Figure / Table
% \renewcommand{\figurename}{Fig.}
% \renewcommand{\tablename}{Table}

\setcounter{figure}{0}
\setcounter{table}{0}
\counterwithin{figure}{section}
% \section{Traffic Simulation Environment}  % 附录A
% \subsection{Heterogeneous Intersections and Signal Phases} \label{HI}
\section{Intersection Modeling}
\begin{figure}[h]
  \centering

  % ===== 可调参数：0–1 表示列宽比例 =====
  % 左图：左下角(\Lx,\Ly)，宽\Lw，高\Lh
  \newcommand{\Lx}{0.05}
  \newcommand{\Ly}{0.02}
  \newcommand{\Lw}{0.5}
  \newcommand{\Lh}{0.5}

  % 右上图：左下角(\Rax,\Ray)，宽\Raw，高\Rah
  \newcommand{\Rax}{0.6}
  \newcommand{\Ray}{0.35}
  \newcommand{\Raw}{0.3}
  \newcommand{\Rah}{0.3}

  % 右下图：左下角(\Rbx,\Rby)，宽\Rbw，高\Rbh
  \newcommand{\Rbx}{0.6}
  \newcommand{\Rby}{0.02}
  \newcommand{\Rbw}{0.3}
  \newcommand{\Rbh}{0.3}

  % 标注与辅助
  \newcommand{\capgap}{2pt}        % 小标题与图的垂直间隔
  \newif\ifshowframe \showframefalse % 设为 \showframetrue 可显示调试边框

  \begin{tikzpicture}[x=\columnwidth, y=\columnwidth]
    % 保留版面空间（0,0 左下；1,1 右上）
   \path[use as bounding box] (0,0) rectangle (1,0.45);

    % ----- 左图 -----
    \node[anchor=south west, inner sep=0pt] (Limg) at (\Lx,\Ly)
      {\includegraphics[width=\Lw\columnwidth,height=\Lh\columnwidth,keepaspectratio]{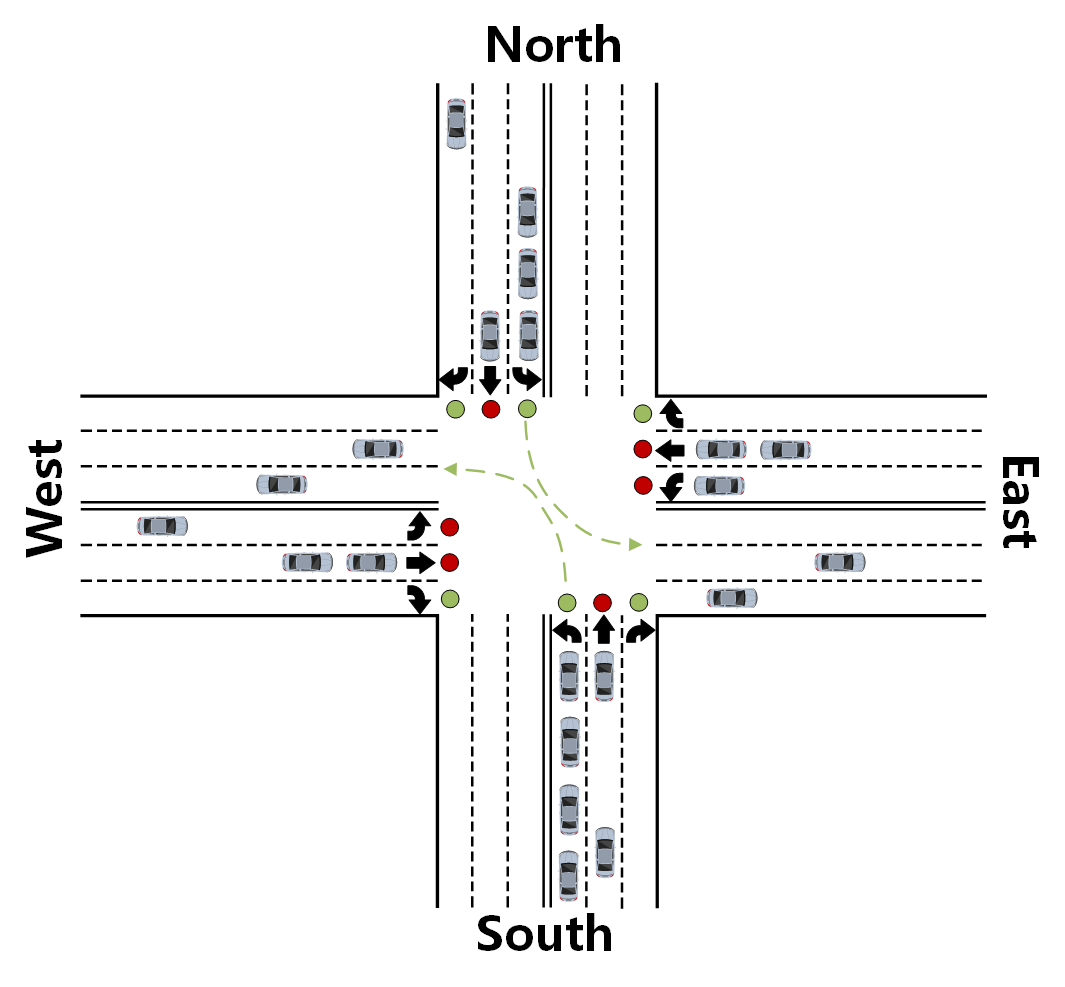}};
    \ifshowframe\draw[gray] (\Lx,\Ly) rectangle ++(\Lw,\Lh);\fi
    \node[anchor=north west, yshift=-\capgap, xshift=40pt] at (Limg.south west) {\scriptsize (a) Intersection};

    % ----- 右上 -----
    \node[anchor=south west, inner sep=0pt] (R1img) at (\Rax,\Ray)
      {\includegraphics[width=\Raw\columnwidth,height=\Rah\columnwidth,keepaspectratio]{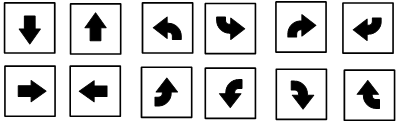}};
    \ifshowframe\draw[gray] (\Rax,\Ray) rectangle ++(\Raw,\Rah);\fi
    \node[anchor=north west, yshift=-\capgap, xshift=9pt] at (R1img.south west) {\scriptsize (b) Twelve Lanes};

    % ----- 右下 -----
    \node[anchor=south west, inner sep=0pt] (R2img) at (\Rbx,\Rby)
      {\includegraphics[width=\Rbw\columnwidth,height=\Rbh\columnwidth,keepaspectratio]{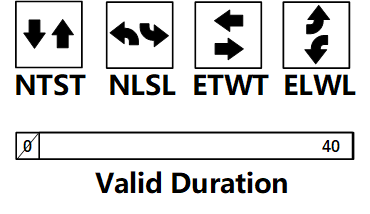}};
    \ifshowframe\draw[gray] (\Rbx,\Rby) rectangle ++(\Rbw,\Rbh);\fi
    \node[anchor=north west, yshift=-\capgap, xshift=15pt] at (R2img.south west) {\scriptsize (c) Actions};
  \end{tikzpicture}

  \caption{Intersection Modeling}
  \label{fig:Intersection}
\end{figure}

\bibliographystyle{IEEEtran}
\bibliography{ref}

% Generated by IEEEtran.bst, version: 1.14 (2015/08/26)
\begin{thebibliography}{10}
\providecommand{\url}[1]{#1}
\csname url@samestyle\endcsname
\providecommand{\newblock}{\relax}
\providecommand{\bibinfo}[2]{#2}
\providecommand{\BIBentrySTDinterwordspacing}{\spaceskip=0pt\relax}
\providecommand{\BIBentryALTinterwordstretchfactor}{4}
\providecommand{\BIBentryALTinterwordspacing}{\spaceskip=\fontdimen2\font plus
\BIBentryALTinterwordstretchfactor\fontdimen3\font minus \fontdimen4\font\relax}
\providecommand{\BIBforeignlanguage}[2]{{%
\expandafter\ifx\csname l@#1\endcsname\relax
\typeout{** WARNING: IEEEtran.bst: No hyphenation pattern has been}%
\typeout{** loaded for the language `#1'. Using the pattern for}%
\typeout{** the default language instead.}%
\else
\language=\csname l@#1\endcsname
\fi
#2}}
\providecommand{\BIBdecl}{\relax}
\BIBdecl

\bibitem{guo2023urban}
J.~Guo, S.~Ghanadbashi, S.~Wang, and F.~Golpayegani, ``Urban traffic signal control at the edge: An ontology-enhanced deep reinforcement learning approach,'' in \emph{2023 IEEE 26th International Conference on Intelligent Transportation Systems (ITSC)}.\hskip 1em plus 0.5em minus 0.4em\relax IEEE, 2023, pp. 6027--6033.

\bibitem{xu2021hierarchically}
B.~Xu, Y.~Wang, Z.~Wang, H.~Jia, and Z.~Lu, ``Hierarchically and cooperatively learning traffic signal control,'' in \emph{Proceedings of the AAAI conference on artificial intelligence}, vol.~35, no.~1, 2021, pp. 669--677.

\bibitem{gu2024pi}
Y.~Gu, K.~Zhang, Q.~Liu, W.~Gao, L.~Li, and J.~Zhou, ``$\pi$-light: Programmatic interpretable reinforcement learning for resource-limited traffic signal control,'' in \emph{Proceedings of the AAAI Conference on Artificial Intelligence}, vol.~38, no.~19, 2024, pp. 21\,107--21\,115.

\bibitem{liang2022oam}
E.~Liang, Z.~Su, C.~Fang, and R.~Zhong, ``{OAM}: An option-action reinforcement learning framework for universal multi-intersection control,'' in \emph{Proceedings of the AAAI Conference on Artificial Intelligence}, vol.~36, no.~4, 2022, pp. 4550--4558.

\bibitem{lai2023large}
S.~Lai, Z.~Xu, W.~Zhang, H.~Liu, and H.~Xiong, ``Large language models as traffic signal control agents: Capacity and opportunity,'' \emph{arXiv preprint arXiv:2312.16044}, 2023.

\bibitem{10763740}
J.~Fan, H.~Chu, L.~Liu, and H.~Ma, ``Llmair: Adaptive reprogramming large language model for air quality prediction,'' in \emph{2024 IEEE 30th International Conference on Parallel and Distributed Systems (ICPADS)}, 2024, pp. 423--430.

\bibitem{10763542}
X.~Chen, J.~Cumin, F.~Ramparany, and D.~Vaufreydaz, ``Towards llm-powered ambient sensor based multi-person human activity recognition,'' in \emph{2024 IEEE 30th International Conference on Parallel and Distributed Systems (ICPADS)}, 2024, pp. 609--616.

\bibitem{da2024prompt}
L.~Da, M.~Gao, H.~Mei, and H.~Wei, ``Prompt to transfer: Sim-to-real transfer for traffic signal control with prompt learning,'' in \emph{Proceedings of the AAAI Conference on Artificial Intelligence}, vol.~38, no.~1, 2024, pp. 82--90.

\bibitem{da2024open}
L.~Da, K.~Liou, T.~Chen, X.~Zhou, X.~Luo, Y.~Yang, and H.~Wei, ``Open-ti: Open traffic intelligence with augmented language model,'' \emph{International Journal of Machine Learning and Cybernetics}, pp. 1--26, 2024.

\bibitem{pang2024illm}
A.~Pang, M.~Wang, M.-O. Pun, C.~S. Chen, and X.~Xiong, ``illm-tsc: Integration reinforcement learning and large language model for traffic signal control policy improvement,'' \emph{arXiv preprint arXiv:2407.06025}, 2024.

\bibitem{10.1145/3703155}
\BIBentryALTinterwordspacing
L.~Huang, W.~Yu, W.~Ma, W.~Zhong, Z.~Feng, H.~Wang, Q.~Chen, W.~Peng, X.~Feng, B.~Qin, and T.~Liu, ``A survey on hallucination in large language models: Principles, taxonomy, challenges, and open questions,'' \emph{ACM Trans. Inf. Syst.}, vol.~43, no.~2, Jan. 2025. [Online]. Available: \url{https://doi.org/10.1145/3703155}
\BIBentrySTDinterwordspacing

\bibitem{10535743}
M.~Wang, X.~Xiong, Y.~Kan, C.~Xu, and M.-O. Pun, ``Unitsa: A universal reinforcement learning framework for v2x traffic signal control,'' \emph{IEEE Transactions on Vehicular Technology}, vol.~73, no.~10, pp. 14\,354--14\,369, 2024.

\bibitem{zhang2024dynamiclighttwostagedynamictraffic}
\BIBentryALTinterwordspacing
L.~Zhang, Y.~Zhang, S.~Xie, J.~Deng, and C.~Li, ``{DynamicLight}: Two-stage dynamic traffic signal timing,'' 2024. [Online]. Available: \url{https://arxiv.org/abs/2211.01025}
\BIBentrySTDinterwordspacing

\bibitem{kim2023prioritized}
H.~Kim, Z.~Jin, H.~Tak, H.~Yu, and H.~Yeo, ``Prioritized phase split optimization for coordinated traffic signal control in urban network using deep reinforcement learning,'' in \emph{2023 IEEE 26th International Conference on Intelligent Transportation Systems (ITSC)}.\hskip 1em plus 0.5em minus 0.4em\relax IEEE, 2023, pp. 833--838.

\bibitem{10763793}
W.~Dong, W.~Liu, R.~Xi, M.~Hou, and S.~Fan, ``Mletune: Streamlining database knob tuning via multi-llms experts guided deep reinforcement learning,'' in \emph{2024 IEEE 30th International Conference on Parallel and Distributed Systems (ICPADS)}, 2024, pp. 226--235.

\bibitem{hu2022lora}
E.~J. Hu, Y.~Shen, P.~Wallis, Z.~Allen-Zhu, Y.~Li, S.~Wang, L.~Wang, and W.~Chen, ``Lo{RA}: Low-rank adaptation of large language models,'' in \emph{International Conference on Learning Representations}, 2022.

\bibitem{wang2024making}
P.~Wang, L.~Li, L.~Chen, F.~Song, B.~Lin, Y.~Cao, T.~Liu, and Z.~Sui, ``Making large language models better reasoners with alignment,'' 2024.

\bibitem{wei2019survey}
H.~Wei, G.~Zheng, V.~Gayah, and Z.~Li, ``A survey on traffic signal control methods,'' \emph{arXiv preprint arXiv:1904.08117}, 2019.

\bibitem{Zhang2019CityFlowAM}
\BIBentryALTinterwordspacing
H.~Zhang, S.~Feng, C.~Liu, Y.~Ding, Y.~Zhu, Z.~Zhou, W.~Zhang, Y.~Yu, H.~Jin, and Z.~J. Li, ``{CityFlow}: A multi-agent reinforcement learning environment for large scale city traffic scenario,'' \emph{The World Wide Web Conference}, 2019. [Online]. Available: \url{https://api.semanticscholar.org/CorpusID:153312553}
\BIBentrySTDinterwordspacing

\bibitem{Koonce2008}
P.~Koonce, ``Traffic signal timing manual,'' Tech Report FHWA-HOP-08-024, 2008.

\bibitem{VARAIYA2013177}
P.~Varaiya, ``Max pressure control of a network of signalized intersections,'' \emph{Transportation Research Part C: Emerging Technologies}, vol.~36, pp. 177--195, 2013.

\bibitem{10.1145/3292500.3330949}
H.~Wei, C.~Chen, G.~Zheng, K.~Wu, V.~Gayah, K.~Xu, and Z.~Li, ``Presslight: Learning max pressure control to coordinate traffic signals in arterial network,'' in \emph{Proceedings of the 25th ACM SIGKDD International Conference on Knowledge Discovery \& Data Mining}, ser. KDD '19.\hskip 1em plus 0.5em minus 0.4em\relax New York, NY, USA: Association for Computing Machinery, 2019, p. 1290–1298.

\bibitem{Chen_Wei_Xu_Zheng_Yang_Xiong_Xu_Li_2020}
C.~Chen, H.~Wei, N.~Xu, G.~Zheng, M.~Yang, Y.~Xiong, K.~Xu, and Z.~Li, ``Toward a thousand lights: Decentralized deep reinforcement learning for large-scale traffic signal control,'' \emph{Proceedings of the AAAI Conference on Artificial Intelligence}, vol.~34, no.~04, pp. 3414--3421, Apr. 2020.

\bibitem{10.1145/3357384.3357902}
H.~Wei, N.~Xu, H.~Zhang, G.~Zheng, X.~Zang, C.~Chen, W.~Zhang, Y.~Zhu, K.~Xu, and Z.~Li, ``Colight: Learning network-level cooperation for traffic signal control,'' in \emph{Proceedings of the 28th ACM International Conference on Information and Knowledge Management}, ser. CIKM '19.\hskip 1em plus 0.5em minus 0.4em\relax New York, NY, USA: Association for Computing Machinery, 2019, p. 1913–1922.

\bibitem{Wu2021EfficientPI}
Q.~Wu, L.~Zhang, J.~Shen, L.~Lu, B.~Du, and J.~Wu, ``Efficient pressure: Improving efficiency for signalized intersections,'' \emph{ArXiv}, vol. abs/2112.02336, 2021.

\bibitem{advanced_xlight}
L.~Zhang, Q.~Wu, J.~Shen, L.~L{\"u}, B.~Du, and J.~Wu, ``Expression might be enough: representing pressure and demand for reinforcement learning based traffic signal control,'' in \emph{International Conference on Machine Learning}.\hskip 1em plus 0.5em minus 0.4em\relax PMLR, 2022, pp. 26\,645--26\,654.

\bibitem{NEURIPS2020_29e48b79}
A.~Oroojlooy, M.~Nazari, D.~Hajinezhad, and J.~Silva, ``Attendlight: Universal attention-based reinforcement learning model for traffic signal control,'' in \emph{Advances in Neural Information Processing Systems}, H.~Larochelle, M.~Ranzato, R.~Hadsell, M.~Balcan, and H.~Lin, Eds., vol.~33.\hskip 1em plus 0.5em minus 0.4em\relax Curran Associates, Inc., 2020, pp. 4079--4090.

\bibitem{zou2025trafficr1reinforcedllmsbring}
\BIBentryALTinterwordspacing
X.~Zou, Y.~Yang, Z.~Chen, X.~Hao, Y.~Chen, C.~Huang, and Y.~Liang, ``Traffic-r1: Reinforced llms bring human-like reasoning to traffic signal control systems,'' 2025. [Online]. Available: \url{https://arxiv.org/abs/2508.02344}
\BIBentrySTDinterwordspacing

\end{thebibliography}

\end{document}